\documentclass{article}
\usepackage{amsmath}
\usepackage{algorithm}
\usepackage{algpseudocode}
\usepackage{amssymb}
\usepackage[most]{tcolorbox}
\usepackage{xcolor}
\usepackage{adjustbox}
\usepackage{setspace}

\definecolor{lightblue}{RGB}{230,245,255}
\definecolor{lightred}{RGB}{255,230,230}
\definecolor{darkgreen}{RGB}{0,100,0}
\tcbuselibrary{listingsutf8}
\newtcolorbox[auto counter, number within=section]{functionbox}[2][]{colback=lightblue!10!white, colframe=blue!75!black, fonttitle=\bfseries, title=Function: #2,#1}
\newtcolorbox[auto counter, number within=section]{functionbox2}[2][]{colback=lightgreen!10!white, colframe=green!75!black, fonttitle=\bfseries, title=Function: #2,#1}

\tcbset{enhanced}

\usepackage[preprint]{neurips_2024}

\usepackage[utf8]{inputenc} 
\usepackage[T1]{fontenc}    
\usepackage{hyperref}       
\usepackage{url}            
\usepackage{booktabs}       
\usepackage{amsfonts}       
\usepackage{nicefrac}       
\usepackage{microtype}      
\usepackage{xcolor}         
\usepackage{algorithmicx} 
\usepackage{graphicx} 

\title{Incentivizing Permissionless Distributed Learning of LLMs}

\author{%
  \textbf{Joel Lidin}\textsuperscript{1}
  \thanks{Correspondence to \texttt{contact@tplr.ai}, code can be found at \texttt{tplr.ai}}
  \quad
  \textbf{Amir Sarfi}\textsuperscript{1} \quad
  \textbf{Evangelos Pappas}\textsuperscript{1} \quad
  \textbf{Samuel Dare}\textsuperscript{1} \\
  \textbf{Eugene Belilovsky}\textsuperscript{2} \quad
  \textbf{Jacob Steeves}\textsuperscript{3} \\
  \textsuperscript{1}Templar AI \quad
  \textsuperscript{2}Concordia University, Mila \quad
  \textsuperscript{3}Opentensor Foundation \\
}

\begin{document}

\maketitle

\begin{abstract}
  We describe an incentive system for distributed deep learning of foundational models where peers are rewarded for contributions. The incentive system, \textit{Gauntlet}, has been deployed on the bittensor blockchain and used to train a 1.2B LLM with completely permissionless contributions of pseudo-gradients: no control over the users that can register or their hardware. \textit{Gauntlet} can be applied to any synchronous distributed training scheme that relies on aggregating updates or pseudo-gradients. We rely on a two-stage mechanism for fast filtering of peer uptime, reliability, and synchronization, combined with the core component that estimates the loss before and after individual pseudo-gradient contributions. We utilized an OpenSkill rating system to track competitiveness of pseudo-gradient scores across time. Finally, we introduce a novel mechanism to ensure peers on the network perform unique computations. Our live 1.2B run, which has paid out real-valued tokens to participants based on the value of their contributions, yielded a competitive (on a per-iteration basis) 1.2B model that demonstrates the utility of our incentive system.   
\end{abstract}

\section{Introduction}



Training large foundational models, such as large language models (LLMs), remains dominated by centralized actors with access to vast computational resources. However, as these models grow in importance and influence, so does the imperative to democratize their development. 
Foundational models such as LLMs are typically trained in large, well-interconnected centralized data centers, in part due to the heavy communication costs of typical data parallel distributed learning methods. However, recent advancements in communication-efficient distributed learning \cite{wang2023cocktailsgd,douillard2023diloco,peng2024decoupled,ahn2025dion} open the possibility to decentralize computation. This opens the door to new paradigms such as permissionless distributed training, where anyone can contribute updates to a shared model. While recent works have demonstrated pre-training with decentralized or federated learning\cite{borzunov2022training,jaghouar2024intellect,sani2024future}, these approaches have not fully addressed the challenges of incentivization and quality control in open networks opting for vetted contributors. In particular, ensuring honest user participation in such systems remains a challenge. 

Recent work has considered verification of machine learning programs\cite{arun2025verde}, from untrusted users which can include distributed training. However, this work on verification focuses on assuring that exact pre-described computations have been performed. On the other hand, contributors in incentivized systems may vary their computations (e.g., data selection or hyperparameters). Proof of Learning (PoL) has also been proposed in \cite{jia2021proof} and can be seen as a type of verification mechanism that attempts to recompute the exact calculations that a learner should have performed. It avoids re-training of the entire model by only recomputing part of the computation. Similar to verification, existing PoL work does not consider the case where a learner can deviate and even improve on the prescribed learning scheme (such as using more data than specified for a gradient step).

Incentivized distributed training can be seen as a generalization of verification systems in distributed learning. The goal is to both ensure that untrusted users provide useful computations to the system, and to incentivize competition and innovation between users. For example, the incentive system may encourage participants to optimize their local hardware, networking, as well as their implementation in order to maximize throughput and utility of their contributions.

In this paper, we introduce the Gauntlet incentive mechanism, a system designed to enable and reward high-quality contributions in a permissionless distributed training setting. Gauntlet efficiently evaluates and compares pseudo-gradient contributions from peers participating in a distributed training run. Gauntlet has been deployed on the Bittensor blockchain and used in a live training run of a 1.2B parameter language model, where contributors provided compressed pseudo-gradients without any centralized registration or approval. This model achieved competitive performance per iteration, with the Gauntlet protocol paying real-valued tokens to participants in proportion to the utility of their updates.


Our 1.2B model run is, to our knowledge, the first truly permissionless pre-training LLM run. Any user with a valid internet connection is able to make a contribution without needing approval, coordination, or identification. It demonstrates that foundational model training can be conducted in a completely open network with minimal assumptions about trust, identity, or compute capabilities. Our study opens the door to decentralized AI models sustained by market-driven incentives.






\section{Distributed Training Framework}
We consider a general synchronous distributed training scheme where a model $\theta$ is updated with learning rate $\alpha$ as follows: 

\begin{equation}\label{eq:aggregation} 
    \theta_t = \theta_{t-1} - \alpha\sum_{k=1}^Kw_k{\Delta_k}
\end{equation}

In this scheme, $K$ distributed peers contribute "pseudo-gradients", $\Delta_k$, which are aggregated and used to perform an optimization step. We note that this is a generic framework encompassing a number of popular data parallel distributed learning schemes \cite{goyal2017accurate, reddi2020adaptive, qian2021error, douillard2023diloco, wang2023cocktailsgd,shi2019understanding, peng2024decoupled}

Training proceeds in communication rounds $t$, each with a specified duration. 
At the end of each round, we define a `put window': a short period during which peers must publish their pseudo-gradients. Submissions made outside this window (i.e., too early or too late) are ignored.

To implement incentivization, we consider the concept of a validator that can access any contribution $\Delta_k$ for evaluation. The validator maintains a score for all participants, which is periodically posted on the public blockchain. This score is used to determine the amount of monetary rewards given to participants on the network. 

The instantiation of this framework that we evaluate uses the communication-efficient Decoupled Momentum Optimizer (DeMo)\cite{peng2024decoupled} to produce pseudo-gradients (see Algo. 2). DeMo is a variant of compression with error feedback methods \cite{karimireddy2019error,shi2019understanding,wang2023cocktailsgd}. The compressor utilized by DeMo applies a Discrete Cosine Transform (DCT) operation on chunked tensors, decorrelating the values before applying a top-k operation. This method has been shown to achieve competitive compression ratios on LLM training \cite{peng2024decoupled,ahn2025dion}. 


\section{Gauntlet Incentive} 
Our incentive system is built with two phases: (a) a compute-intensive primary evaluation applied to a small number of peers per communication round (b) a low-cost fast evaluation applied to a large number of peers in each round. Incentives are calculated in each communication window and the scores of each participant are updated locally by the validator. The overall behavior of peers and validators in the system is summarized in Algorithm 1.

\subsection{Primary evaluation}

\paragraph{Loss Rating}
The heart of the incentive mechanism attempts to judge the value of each pseudo-gradient contribution.

\begin{equation}
\text{LossScore}_p(\Delta_t^p,D) = L(\theta_t,D)- L(\theta_t-\beta\Delta_t^p,D)
\end{equation}

where $\Delta_t^p$ is the pseudo-gradient from peer $p$ at round $t$, $D$ is a random subset of data from the training dataset, and $\beta$ is a scaling factor. This essentially measures how much a peer's contribution decreases the loss. Naturally, poor gradients will lead to highly negative scores allowing the system to quickly downweight malicious contributors. 

Note that since an individual contribution has a higher variance than the aggregated pseudo-gradients, the $\beta$ will typically be set to a smaller value than the current learning rate. In practice, when using a learning rate scheduler, we found it was sufficient to set $\beta_t = c*\alpha_t$ where $c<1$. Using a smaller value also allows us to reduce the noise in the $\text{LossScore}$. Specifically, stepping with too large a step size is more likely to lead to negative loss scores, and in our empirical observations inconsistent rankings between peers.

A significant issue with loss-based scores is that they are not consistent over time; indeed, even adjacent iterates can lead to very different scores for the same peer running the same strategy. This problem is exacerbated by the fact that practically, the validator cannot evaluate all peers' contributions at each communication round. On the other hand, we observed that at any given round, ranking based on LossScores correlated well with high quality contributions (e.g. peers processing more data achieved better LossScores). We thus utilize a rank-based rating system OpenSkill \cite{joshy2024openskill}, which is well suited to estimating relative peer ranks under sparse evaluation. 

In each evaluation round $t$, a random subset $S$ of the $K$ participating peers is chosen and ranked by their $\text{LossScore}_p$. Subsequently their OpenSkill rating, $\text{LossRating}_p$, is updated.

\paragraph{Proof of Computation} A key challenge in a completely open permissionless system is that peers broadcast their pseudo-gradients to all peers on the network. This leads to several related problems where peers can avoid performing computation while achieving positive loss scores:

\begin{itemize}
    \item Peer Copying  - A peer attempts to copy a valid pseudo-gradient uploaded by another peer and post it before the communication period is completed.%
    \item Duplicating Contributions - A peer attempts to register multiple times on the network and upload identical pseudo-gradients
\end{itemize}

Our proposed solution relies on assigning a unique subset of data $D_t^p$ to peer $p$ at any given round that must be used as part of its training data for that round. The validator then attempts to determine if the peer has actually performed training on this data by comparing the loss score on this data to the loss score on a random subset of data (already computed as part of the Loss Rating). 

\begin{equation}
   \mu_p = \gamma\mu_p +(1-\gamma)*\operatorname{sign}(\text{LossScore}_p(\Delta_t^p,D_t^p) - \text{LossScore}_p(\Delta_t^p,D_t^{rand}))
\end{equation}

Peers training on their assigned data $D_t^p$ are expected to have lower loss on this data compared to their loss on a random data subset $D_t^{rand}$. This difference tends to yield $\mu_p>0$ over time for compliant peers. Conversely, peers neglecting $D_t^p$ are expected to have $\mu_p \simeq 0$. The resulting $\mu_p$ contributes to the peer's overall incentivization score, as detailed in equation~\ref{eq:alltogether}.



\paragraph{Signed Descent} Following \cite{peng2024decoupled}, we utilize the $\operatorname{sign}$ operation post-aggregation, which provides a number of practical benefits: (a) gradient norm control (b) ability to store the aggregation to allow fast checkpoint catchup. Specifically checkpointing can occur infrequently while catchup can be done through repeated application of the signed updates. For consistency the use of the sign is also done at evaluation.

\subsection{Fast Evaluation}
On a larger subset of peers we perform the following low cost evaluations including basic sanity checks and a score to estimate synchronization of the local model with the expected model on the validator.

\paragraph{Basic checks} We penalize peers for the following: (a) not sending their pseudo-gradient within the specified put window. This is facilitated by our use of cloud-based storage in combination with the blockchain time (detailed in the next section), which provides a consistent global clock. (b) Not putting a pseudo-gradient at all (c) violating the format (e.g., submitting tensors with incorrect dimensions or data types).

\paragraph{Sync Score} In each communication round, peers also send a very small number of their model parameters (2 values per tensor). The cost of this is negligible compared to the overall communication cost. From this the validator computes a synchronization score as follows from the $N$ communicated parameters: 
\[
\textsc{SyncScore} = \frac{1}{\alpha N} \sum_{i=1}^N \left| \theta_i^{(validator)} - \theta_i^{(peer)} \right|
\]

Given that pseudo-gradient updates are signed post-aggregation (effectively quantizing updates by the learning rate $\alpha$), the Sync Score provides a heuristic measure of how many update steps a peer's model diverges from the validator's. We use a threshold for this score as a filter (in practice, setting the threshold to 3).

Violation of either the basic checks or the sync score constraint leads to an additional penalty:
\[
\phi_p =
\begin{cases}
    0.75 & \text{if peer } p \text{ fails any fast evaluation check} \\
    1 & \text{otherwise}
\end{cases}
\]

We apply the penalty directly to $\mu_p$ each time fast evaluation is performed
\[\mu_p = \phi*\mu_p\]
This allows to rapidly degrade the score when a peer repeatedly fails the fast evaluation, as will be discussed in the next section. It also allows the peer to be quickly removed from the aggregation. 

\begin{algorithm}
\caption{Gauntlet Incentive Scheme}
\begin{algorithmic}
\setstretch{.2}
\Require Number of peers $K$, iterations $T$, learning rate $\alpha$, EMA decay $\beta$
\Statex Initialize model parameters $\theta_0$
\Statex Initialize generalization scores $\mu_p \gets 0$ for all $p$
\Statex Initialize peer scores $\textsc{PeerScore}_p \gets 0$ for all $p$

    \setlength{\parskip}{2mm}
    \Statex
    \begin{tcolorbox}[colback=lightblue, colframe=blue!50, title=\textbf{Peers}] \vspace{-1mm}
    \For{$t = 0$ to $T-1$}
    \For{each peer $p \in \{1, \dots, K\}$ \textbf{in parallel}}
        \State $D_t^p \gets \textsc{SelectData}(seed, p, t)$
        \State $\Delta_t^p \gets \textsc{PseudoGradient}(D_t^p, \theta_t)$
        \State $\textsc{Broadcast}(\Delta_t^p)$
        \Statex \hspace{3em} \textcolor{darkgreen}{// Local aggregation and update }
        \State $w \gets \textsc{SelectTopG}(\{\textsc{PeerScore}_w\}_{w=1}^K, G)$
        \State $\Delta_t^{\text{agg}} \gets \textsc{Aggregate}(\{\Delta_t^w \mid w_p > 0\})$

        \State $\theta_{t+1}^{p} \gets \theta_t - \alpha \Delta^{\text{agg}}_t$
    \EndFor
    \EndFor
    \end{tcolorbox}
\Statex
\begin{tcolorbox}[colback=lightred, colframe=red!50, title=\textbf{Validator}] \vspace{-1mm}
\For{$t = 0$ to $T-1$} \vspace{1mm}
    \State \textcolor{darkgreen}{// Evaluate a small set of peers}\vspace{.1mm}
    \State $S_t \gets \textsc{SelectRandomPeers}(v)$ \Comment{Evaluation set, $|S_t| \ll K$}
    \Statex

        \State \textcolor{darkgreen}{// Update scores for a larger set using filtering}
    \State $F_t \gets \textsc{SelectPeersForFiltering}(F)$ 
    
    \For{each peer $p \in F_t$}
        \State $\phi_p \gets \textsc{Fast Evaluation}(p)$
        \Comment{Check if Peer Passes Fast Evaluation}
        \State $\mu_p \gets \phi_p\cdot \mu_p$ \vspace{0.9mm}
    \EndFor
    \Statex \vspace{2mm}
    \For{each peer $p \in S_t$}
        \Statex \hspace{3em} \textcolor{darkgreen}{// Evaluate on peer's assigned data}
        \State $\theta'_p \gets \theta_t - \alpha \textsc{Sign}(\Delta_t^p)$
    
        \State $D_t^p \gets \textsc{SelectData}(seed, p, t)$
        \State $\delta^{\text{assigned}}_p \gets L(\theta_t, D_t^p) - L(\theta'_p, D_t^p)$\vspace{0.2mm}
    
        \Statex \hspace{3em} \textcolor{darkgreen}{// Evaluate on unassigned (random) data}\vspace{0.2mm}
        \State $D_t^{\text{rand}} \gets \textsc{UnassignedData}(p, t)$\vspace{0.9mm}
        \State $\delta^{\text{rand}}_p \gets L(\theta_t, D_t^{\text{rand}}) - L(\theta'_p, D_t^{\text{rand}})$\vspace{0.9mm}
    \EndFor
    \Statex \vspace{2mm}

    \State \textcolor{darkgreen}{// Rank peers in $S_t$ using generalization signal}
    \State $\{\text{LossRating}_p\}_{p \in S_t} \gets \textsc{OpenSkillMatch}(S_t, \{\delta^{\text{rand}}_p\})$\vspace{1mm}
    
    \For{each peer $p \in S_t$}\vspace{-1mm}
        \State $\mu_p \gets \gamma \cdot \mu_p + (1 - \gamma) \cdot \textsc{sign}(\delta^{\text{assigned}}_p - \delta^{\text{rand}}_p)$
    
        \State $\textsc{PeerScore}_p \gets \text{LossRating}_p \cdot \mu_p$ 
    \EndFor

  \Statex \vspace{2mm}
  \State \hspace{0em} \textcolor{darkgreen}{// Aggregate}
    \State $w \gets \textsc{SelectTopG}(\{\textsc{PeerScore}_w\}_{w=1}^K, G)$
        \State $\Delta_t^{\text{agg}} \gets \textsc{Aggregate}(\{\Delta_t^w \mid w_p > 0\})$
    \State $\theta_{t+1} \gets \theta_t - \alpha \cdot\Delta_t^{\text{agg}}$ \vspace{1.5mm}
\EndFor
\end{tcolorbox}
\end{algorithmic}
\end{algorithm}

\subsection{Putting it all together}
Thus, our final pre-normalized score, $\textsc{PeerScore}$ for peer $p$ in round $t$ is given by:

\begin{equation}\label{eq:alltogether}
    \textsc{PeerScore}_p^t = \mu_p*\text{LossRating}_p 
\end{equation}

Finally, the scores are normalized as follows: 

\begin{equation}
    x^{norm}_p = \frac{(\textsc{PeerScore}_p-\min{\textsc{PeerScore}})^c}{\sum_k(\textsc{PeerScore}_k-\min{\textsc{PeerScore}})^c}
\end{equation}

The validator uses these normalized scores $x^{norm}_p$ to assign incentives (that sum to 1) for all peers. These incentive values $x^{norm}_p$ are posted by the validator to the blockchain and used to determine the monetary reward given by the protocol to each peer. In our current design, we use $c=2$, with the goal to increase competition amongst peers. Indeed, the non-linear incentive is designed to encourage participants to register fewer high-performing peers versus many weaker peers. For example if a user has access to $10$ GPUs it is preferred they take care of optimizing their configuration to produce a single high quality pseudo-gradient  with all $10$ GPU as opposed to registering 10 individual peers.

Finally, the weights $w_p$ are used as part of the weighted aggregation in equation~\ref{eq:aggregation}. For simplicity, in Templar-1B, we choose to set the aggregation weight of peers in the top $G$ to $1/G$ and all others zero. This serves to encourage a smaller number of high quality peers with a large amount of compute behind them while also allowing for natural redundancy as when top peers become less reliable they are quickly swapped out in the aggregation with other high quality peers.

\begin{equation}\label{eq:topkScores}
    w_p^t = 
    \begin{cases}
        1/G & \text{if } x_p^t \in \text{Top-}G(x) \\
        0 & \text{otherwise}
    \end{cases}
\end{equation}

During fast evaluation we ensure that the current top G peers are included in the fast evaluation set, such that they can be rapidly downgraded out of this set (and no longer impact the aggregation) if they begin to fail fast evaluation. 

\paragraph{Coordinated Aggregation} Although peers can freely modify their local implementation, the incentive mechanism pushes peers to perform the same aggregation as specified by the validator (e.g., using the same set of peers $G$ specified by the validator) in order to stay synchronized to the validator state $\theta_t^{validator}$. This allows the system to easily propagate various control mechanisms. For instance, a time window for communication is specified and any pseudo-gradients arriving outside of this time window are ignored by the validator, and thus the peers should also ignore them. Similarly, peers are encouraged to use the peer scores $w_p$ in the same way as the validator (equation~\ref{eq:topkScores}).   

\paragraph{Validator Consensus and Stake} In a decentralized system, the evaluation of incentives must also be decentralized. On the Bittensor blockchain, this is achieved through the use of multiple validators who are required to provide stake—an amount of tokens placed at risk as a form of economic commitment. Validators participate in the evaluation process and are subject to penalties for dishonest or faulty behavior. A set of validators typically operates under the Yuma consensus protocol~\cite{Steeves2022}, which combines the incentives $x_p^{norm}$ from different validators. A full description of the Bittensor validator and consensus mechanisms is beyond the scope of this technical report\footnote{https://docs.bittensor.com/yuma-consensus}. For the sake of simplicity in the current implementation of the protocol, the highest staked validator is chosen to provide the location of consistent checkpoints (for peers joining later or restarting) and the list of top-$G$ peers. However, even these can be decentralized in future iterations.

\begin{algorithm}
\caption{DeMo: PseudoGradient and Aggregation}
\begin{algorithmic}[1]
\Require Current model $\theta_t$, local data batch $D_t^p$, momentum buffer $e_t$, compression hyperparameters $s, k$, error feedback decay $\beta$
\Function{DeMoPseudoGradient}{$\theta_t$, $D_t^p$, $e_t$}
    \State $\tilde{g}_t \gets \textsc{LocalStochasticGradient}(\theta_t, D_t^p)$
    \State $e_t \gets \beta \cdot e_t + \tilde{g}_t$ \Comment{Apply error feedback}
    \State $q_t \gets \textsc{DCTEncode}(e_t)$\Comment{DCT on chuncked tensors}
    \State $\hat{q}_t \gets \textsc{TopKCompress}(q_t, s, k)$
    \State $\hat{z}_t \gets \textsc{DCTDecode}(\hat{q}_t)$
    \State $e_t \gets e_t - \hat{z}_t$ \Comment{Update error feedback}
    \State \Return $q_t$ \Comment{Send compressed pseudo-gradient}
\EndFunction

\Function{DeMoAggregation}{$\{q_t^1, \dots, q_t^G\}$}
    \ForAll{$q_t^k$}
        \State $q_t^k \gets \frac{q_t^k}{\|q_t^k\|_2}$ \Comment{Robustness to individual peers norm}
    \EndFor
    \State $Q_t \gets \textsc{AggregateCompressed}(\{q_t^k\})$ \Comment{Weighted average of updates}
    \State $\Delta_t \gets \textsc{DCTDecode}(Q_t)$ \Comment{Decode aggregated update}
    \State $\Delta_t \gets \textsc{Sign}(\Delta_t)$ \Comment{Apply Signum}
    \State \Return $\Delta_t$
\EndFunction
\end{algorithmic}
\end{algorithm}

\section{Byzantine fault tolerance} A challenge in permissionless systems is that participants can violate the prescribed distributed algorithms either intentionally (e.g. by pseudo-gradient poisoning or rescaling) or due to a fault. This problem in the distributed learning literature is often referred to as byzantine fault tolerance, and problematic participants as byzantine workers \cite{malinovsky2023byzantine,xie2018generalized}. In an incentivized system, peers might also inadvertently provide faulty contributions through good faith attempts to increase their incentives. 

 Our incentive system can quickly reduce the weight of byzantine workers, removing them from the aggregation. For example, peers sending poorly scaled contributions will often receive poor loss scores, and peers that are not synchronized will be downweighted. Despite these measures, the system remains vulnerable to risks such as: (a) peers whose malicious behavior is not detected by the incentive mechanism, and (b) a single bad value sent before the peer can be downweighted. A simple example of (b) is a peer intentionally sending a pseudo-gradient with an excessively large magnitude, enough to disrupt the aggregation if it is included even once. 

Such problems have been studied in the literature on byzantine fault tolerance in federated learning. However, many of the more sophisticated methods for addressing this problem either introduce significant overhead or slow down convergence \cite{xie2018generalized,pillutla2022robust}. Some of the more practical approaches rely on gradient clipping \cite{malinovsky2023byzantine}. 

In Templar-1B, we rely on the sign as in \cite{peng2024decoupled} as a final step in the aggregation, which has been found to help the DeMo method converge for LLMs. 
This naturally reduces the impact of direct attacks on the norm of the final update, but an individual peer can still dominate the aggregation by rescaling their pseudo-gradients. We thus rely on a simple strategy of normalizing the contributions $q_t^k$ (see line 12 Algo 2.) so that each peer contributes equally. As our aggregation is done in the DCT encoded domain, we also perform this normalization on the encoded vectors. 
Since we assume that each participant is training on an i.i.d. subset of the data, we do not anticipate large variations in the norms of valid pseudo-gradients. In practice, we observed that this approach significantly reduced the impact of byzantine peers while having no impact on convergence in the fully cooperative (simulated) setting.

\section{Cloud-Based Communication} 
A novel aspect of our system deployment is the use of cloud-based communication backend. Peers and validators in the network communicate using S3-compliant storage buckets. Each peer on the network creates their own bucket and posts the read-access keys to the blockchain, making them visible to other peers and validators. Broadcasting pseudo-gradients is done by simply writing to a local bucket. This has several advantages:

\begin{itemize}
    \item Peer-to-peer network complexities, such as firewall configuration, are avoided.
    \item Pseudo-gradient contributions can be easily tracked and robustly timestamped.
    \item Cloud providers, such as Cloudflare, have globally distributed networks which can often ensure competitive upload and download times for participants worldwide.
\end{itemize}

A disadvantage is that all communication must pass through the cloud provider, making the system limited to the reliability of the cloud provider. Our incentive mechanism encourages peers to optimize their configuration to work robustly with the cloud provider.

\begin{figure}[t]
    \centering
    \includegraphics[width=\linewidth]{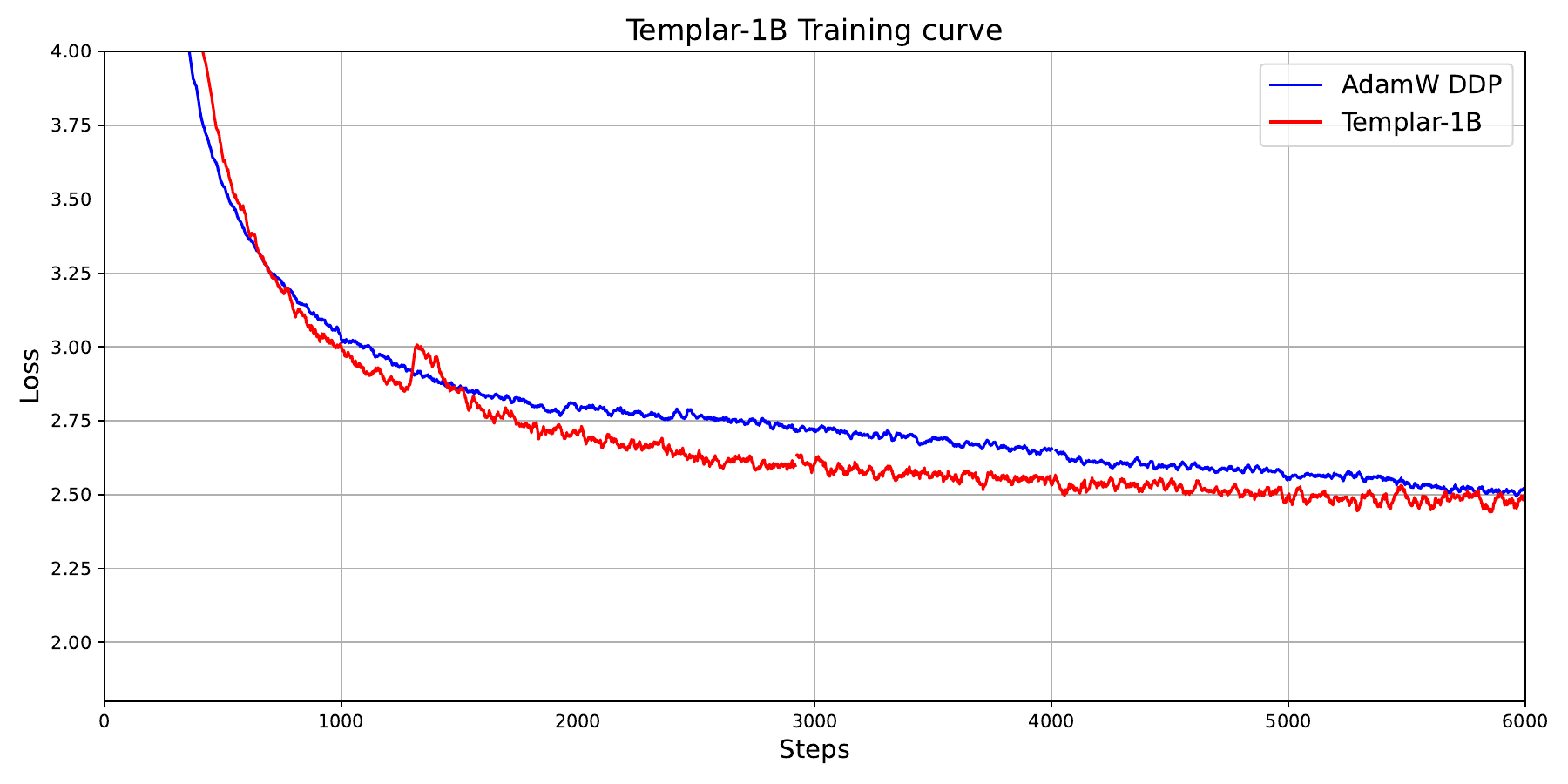}
    \caption{Templar-1B permissionless training curve, compared to a controlled AdamW baseline with the same number of peers and the default per worker batch size. }
    \label{fig:mainfig}
\end{figure}

\section{Results and Discussion}
\begin{figure}[t]
    \centering
    \includegraphics[width=0.8\linewidth]{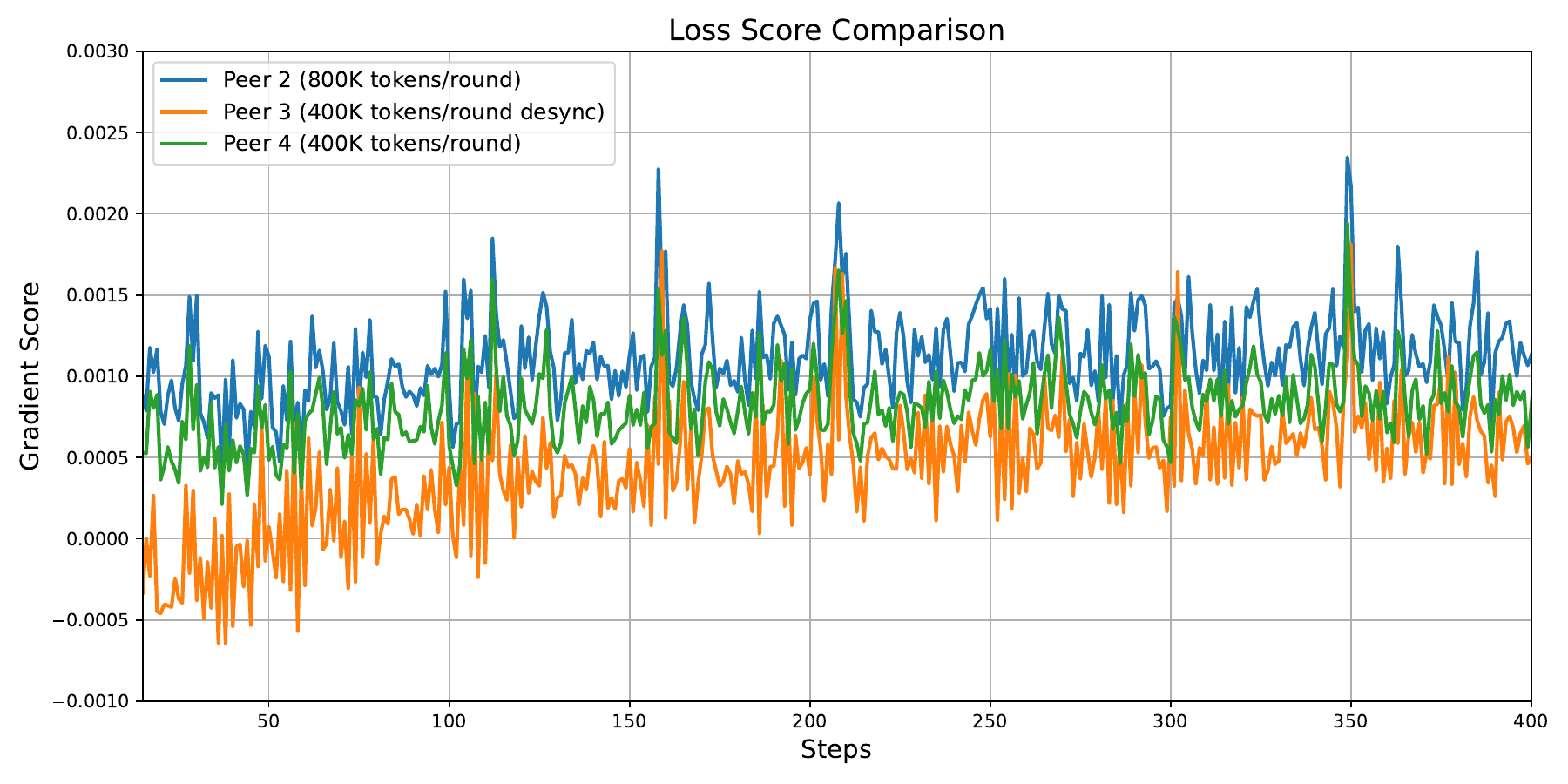}
    \includegraphics[width=0.8\linewidth]{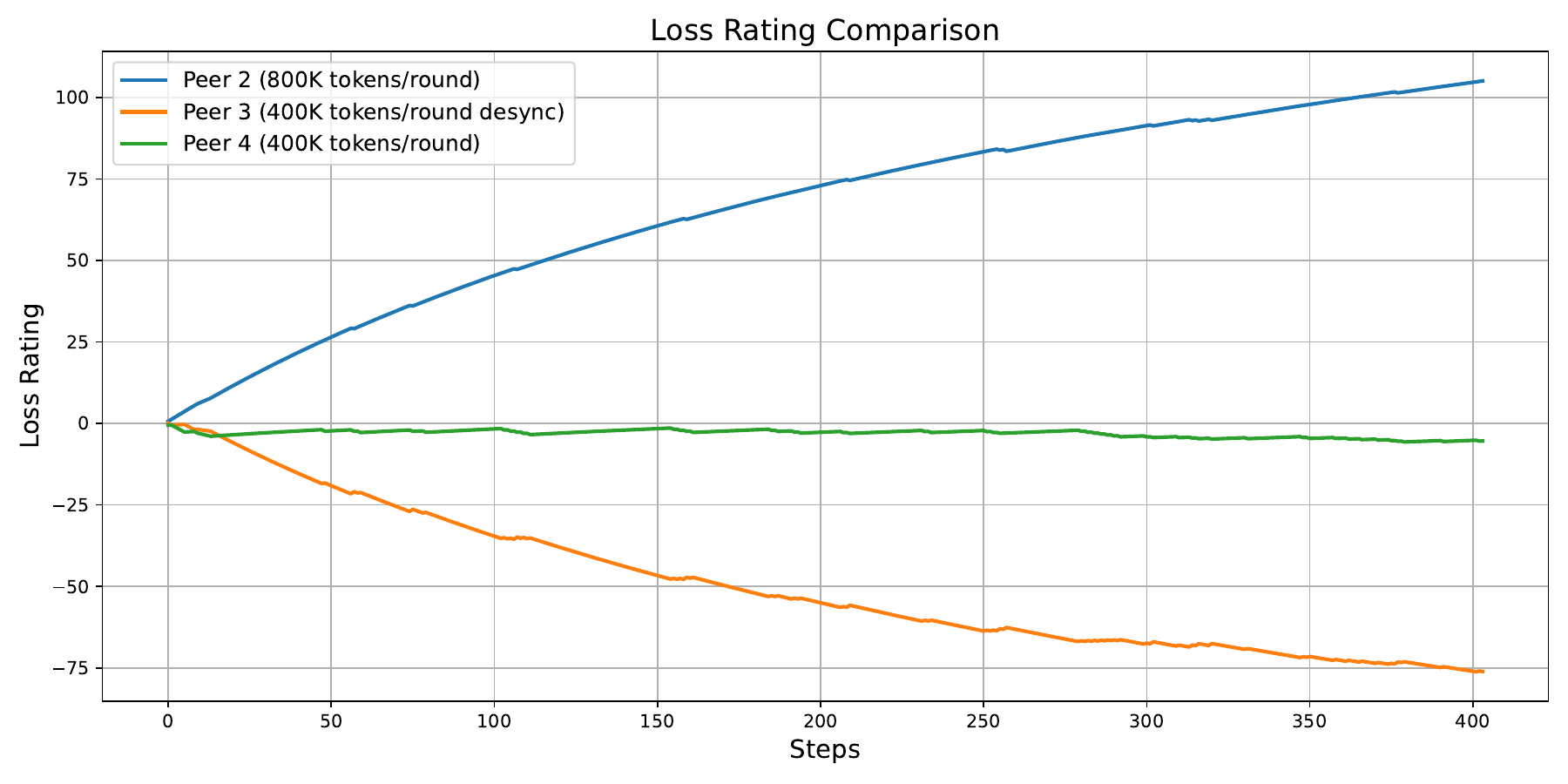}
    \caption{Simulating how LossScore and LossRating evolve for three peers: one processing more data, one desynchronized, and a baseline peer. We observe that the loss score is highly variable from step to step, however relative performance is consistent and the loss rating can quickly differentiate between peers exhibiting favorable behavior.}
    \label{fig:lossrating}
\end{figure}

We deployed Gauntlet to train a 1.2B model for 20K communication rounds in a completely permissionless manner. The evaluation windows were set to be the same as the communication rounds. We used the FineWebEdu dataset\cite{penedo2024fineweb}. Participants were provided a baseline training script, which they could adapt to their particular configuration. The baseline script targeted approximately 400,000 tokens per peer per iteration. However, the length of a communication round was set sufficiently long to allow more data to be processed on a single H100. We aggregated pseudo-gradients from the $G=15$ top-scoring peers in each communication round. The validators were able to evaluate and compare 5 peers each communication round, updating their LossRating and $\mu_p$ score.
The training loss curve is presented in Figure~\ref{fig:mainfig} and compared to an AdamW baseline (with hyperparameters taken from \cite{peng2024decoupled}, training with 15 peers processing 400K tokens per communication round. This represents a comparison to a centralized training algorithm not compatible with training over the internet. We note that, based on prior experiments, the DeMo algorithm roughly follows the convergence dynamics of Adam. Although we cannot measure the exact amount of data each peer processed, we observe that our convergence rate exceeds that of the Adam baseline in the first half of the run, suggesting participants were successfully incentivized to process more data or otherwise optimize their pseudo-gradients (e.g. by tuning local hyperparameters to instantaneously improve loss).

We also compare our downstream metrics against the baseline Adam trained model and the published results of \cite{peng2024decoupled}, which was trained for the same number of iterations in Table~\ref{tab1}. We see that our downstream metrics are competitive. 




\begin{table}[h]
\centering
\caption{Base model evaluation results on downstream benchmarks (zero-shot). We compare with published results of DeMo and custom training with AdamW using the same number of steps. \textsc{Templar-1B} token counts are estimated as number of tokens processed by participants is not controlled.}\label{tab1}
\begin{adjustbox}{max width=\textwidth}
\begin{tabular}{lcccccc}
\toprule
\textbf{Model} & \textbf{Dataset} & \textbf{Tokens} & \textbf{HellaSwag} & \textbf{PIQA} & \textbf{ARC-E} \\
 &  &  &  acc\_norm & acc\_norm & acc \\
\midrule
\textsc{Templar-1B}         & FineWebEdu & 100B-200B & 51.0 & 71.4 & 59.2 \\
\textsc{DeMo} 1B~\cite{peng2024decoupled}        & Dolmo  & 100B       & 48.0 & 70.0 & 55.0 \\
AdamW DDP 1B          & FineWebEdu & 120B       & 51.0   & 71.9   & 58.9    \\
\bottomrule
\end{tabular}
\end{adjustbox}
\end{table}

\paragraph{Simulating LossRating} We performed controlled simulations of the Gauntlet incentive system focusing on identifying whether $\text{LossRating}$ fulfills two basic properties: (a) peers training on more data get higher rating (b) peers who deviate from the global state get downweighted. The desynchronized peer was simulated by having the peer pause early on for 3 communication periods (thus representing a peer who is 3 steps behind) and then continue with the deviating model. In general our simulations showed that the $\text{LossRating}$ can robustly detect both these scenarios, as observed in Figure~\ref{fig:lossrating}. One peer training with 800K tokens per communication round significantly outperforms a peer training with the default 400K tokens per communication round. Similarly, a peer who is delayed by several steps rapidly begins to underperform. 


\paragraph{Synchronous Model States Simplify Validation} Distributed learning methods can be broadly categorized into those that maintain the same model on all peers and those which allow models to diverge (e.g. asynchronous SGD, gossip-based methods). An advantage of methods that allow models to diverge is that they can typically support heterogeneous communication patterns more easily as well as overlapping communication; on the other hand, they are more challenging to debug and work with. We have found that, in the context of incentivization, synchronized model states are critical for allowing the validator to easily compare the contributions of peers. In an earlier experimental version of our system we allowed peers and validators to partially diverge, but this creates significant issues, and differences in evaluation of the loss are challenging to attribute to model divergences between validator and peer states. Furthermore, even in asynchronous methods, peers need to attempt to stay tightly coupled together, thus encouraging synchronization is still important. This, however, is more challenging to do without a single global reference state that can be available for the validator.

\section{Conclusion}
We have introduced an incentive system for distributed permissionless pre-training of LLMs. We demonstrated that our system, when combined with a communication efficient distributed learning scheme, encourages high quality pseudo-gradient contributions from peers participating all over the world. We demonstrated that it can lead to effective convergence of a 1.2B model with completely permissionless peers participating on the bittensor blockchain. 
Future work will consider scaling to larger models and increasing the efficiency of the underlying communication algorithm.
\newpage
\bibliographystyle{plain}  
\bibliography{bibliography}  

\end{document}